\documentclass{article}
\usepackage{spconf,amsmath,amssymb,graphicx}

\usepackage{enumitem}
\usepackage{algorithm}
\usepackage{algpseudocode}
\usepackage{mathtools}
\setlist{nosep, leftmargin=14pt}

\usepackage{todonotes}
\usepackage{textgreek}
\usepackage{hyperref}
\hypersetup{
    colorlinks=true,
    linkcolor=blue,
    citecolor=blue,
    urlcolor=blue,
}
\usepackage{booktabs}
\usepackage{cleveref}

\def\x{{\mathbf x}}

\def\pt{{p_{\theta}}}
\def\q{q_{\phi}}
\def\x{\mathbf x}
\def\hx{\hat{\mathbf x}}
\def\hxp{\widehat{\mathbf{x}^ \prime}}
\def\xp{\mathbf{x}^ \prime}
\def\dice{$\lceil \text{Dice} \rceil$}

\title{Mitigating the Reconstruction-Detection Trade-off in VAE-based Unsupervised Anomaly Detection}

\name{Agathe Senellart $^{1, \star}$ \quad Ma\"elys Solal $^{2, \star}$ \quad St\'ephanie Allassonni\`ere $^{1, \dagger}$ \quad Ninon Burgos $^{2, \dagger}$}

\address{$^1$ Université Paris Cité, Inria, Inserm, HeKA, F-75015 Paris, France \\
    $^2$ Sorbonne Université, Institut du Cerveau - Paris Brain Institute - ICM, CNRS, Inria, Inserm, \\AP-HP, Hôpital de la Pitié Salpêtrière, F-75013, Paris, France \\
    $\star$ / $\dagger$ Denotes shared first / last authorship
}

\begin{document}

%
\maketitle


%
\begin{abstract}

Variational autoencoders are widely used for unsupervised anomaly detection.
Model selection however remains an open question: to remain fully unsupervised, hyper-parameters are often chosen to minimize the reconstruction error on normal samples. In this paper, we reveal a trade-off between reconstruction quality and anomaly detection among $\beta$-VAE models. Models with constrained latent space reach higher detection metrics but lower reconstruction quality. We also assess the performance variability across random seeds and show it is linked to the distance between normal and abnormal latent distributions. From this analysis, we justify and investigate two methods to mitigate the reconstruction-detection trade-off: beta-scheduling and the Sparse VAE. The latter especially shows an improvement in detection while maintaining high reconstruction quality.


\end{abstract}

\section{Introduction}
Unsupervised anomaly detection (UAD) methods based on deep generative models are used to detect rare events, such as lesions in medical images, by learning a model on healthy data only. The limited availability of labeled data motivates the use of UAD methods, along with the potential that these methods enable the detection of various types of anomalies, ranging from image artefacts to lesions related to various pathologies \cite{chen2022unsupervised}.

Among generative models, variational autoencoders (VAE) are widely used to perform UAD at the sample-level, or at the voxel-level, through anomaly maps \cite{baur2021autoencoders,alaverdyan2020regularized}. In both cases, the model is trained on healthy data to learn the healthy distribution. At inference, for a potentially abnormal test sample, anomaly scores can be constructed from the reconstruction errors \cite{cai2024rethinking}, a distance in the latent space \cite{alaverdyan2020regularized}, or both \cite{zhou2022rethinking}.
Voxel-level UAD is more demanding as an anomaly score must be derived for each voxel, to obtain an anomaly map. To do so, autoencoders can be trained on patches instead of the entire image to derive a score for each voxel \cite{alaverdyan2020regularized}. Another approach is to generate a pseudo-healthy version of the abnormal image and to derive an anomaly map from the difference \cite{chen2022unsupervised}. 
Baur et al. \cite{baur2021autoencoders} make the hypothesis that once a VAE is trained on healthy samples, it will reconstruct a pseudo-healthy image when given an abnormal sample as input, by replacing lesions with healthy tissues. 

Defining a proper criterion to select methods and their hyper-parameters in UAD remains an open question. In VAE reconstruction-based methods, models are often selected based on the reconstruction error of healthy inputs. However, recent work shows that models selected at the beginning of the training procedure detect anomalies better than fully trained models, despite a lower reconstruction quality \cite{huijben2025enhancing}. 
But this loss in reconstruction quality is detrimental to the interpretability of the anomaly maps as the pseudo-healthy reconstructions may look quite different from patient data.
This observed trade-off reveals a misalignment between the auxiliary task used for training models (healthy reconstruction) and the inference task (anomaly detection). 

In addition, models with lowest healthy-to-healthy reconstruction error tend to generalize so well that they sometimes reconstruct anomalies in test images \cite{cai2024rethinking}. 
To avoid this issue, some works have identified that a necessary condition for reliable abnormal-to-healthy reconstruction is that the latent codes of abnormal samples belong to the same domain as those of healthy samples \cite{zhou2022rethinking}. 
To attain this objective, several works propose constraining the latent space; for instance by reducing the latent dimension \cite{cai2024rethinking} or by leveraging a cross-entropy loss to further regularize the latent space \cite{zhou2022rethinking}. 
To date, these methods have only been applied to sample-level anomaly detection, and not yet for designing voxel-level anomaly maps.
Adding to the complexity of model selection, a less explored criterion is stability, while recent work has revealed that detection performance can exhibit large variability depending on the random seed \cite{solal2026unsupervised}.


Motivated by these observations, we establish key criteria for voxel-level UAD methods: (i) high-quality reconstruction for interpretability, (ii) strong anomaly detection performance, and (iii) robustness to seed variation. To explore these criteria, we focus on the $\beta$-VAE model as a recent benchmark demonstrated its good performance for detecting dementia-related lesions in FDG PET images \cite{hassanaly2025benchmarking}.
Our contributions are threefold: (1) We explore the impact of constraining the latent space of $\beta$-VAE models with different latent dimensions and $\beta$ values and show there exists a trade-off between image quality and anomaly detection. (2) We study the performance variability across different seeds through an analysis of the latent space. (3) We investigate two methods to mitigate the trade-off: beta-scheduling which implements a progressive regularization during training \cite{fu2019cyclical}, and the Sparse VAE which enforces adaptive dropout on the latent dimensions \cite{antelmi2019sparse}. 

\section{Materials}
\label{sec:materials}


\subsection{Variational Autoencoder}

In the $\beta$-VAE framework \cite{kingma2013autoencoding,higgins2017betavae}, we aim to learn an encoder $\q(z|x)$ to map an observation $x$ from a set of normal observations $\mathcal{X}$ to a lower dimensional variable $z \in \mathbb{R}^d$, and a decoder $\pt(x|z)$ to map latent codes back to observations. Classically, we set $\q(z|x) = \mathcal{N}(\mu(x), \sigma(x))$ and $\pt(x|z) = \mathcal{N}(\mathcal{D}(z),I)$, where $\mu, \sigma, \mathcal{D}$ are neural networks optimized with the loss function 
\small
\begin{equation}
\label{vae_loss}
    \mathcal{L} = \sum_{x \in \mathcal{X}} \mathbb{E}_{z \sim \q(z|x)}\left[\log\pt(x|z)\right] - \beta \text{KL}(\q(z|x)||p(z))\,,
\end{equation}
\normalsize
where the last term refers to the Kullback-Leibler divergence between $\q(z|x)$ and a prior $p(z) = \mathcal{N}(0,I)$. 
The $\beta$ parameter can be increased to encourage the latent codes to lay in a compact space close to $0$. 
In the remaining of the paper, we denote by $\hx = \mathcal{D}(\mu(\x))$ the VAE-reconstruction of an image $\x$.
We use the optimal architectures found in \cite{hassanaly2025benchmarking}.
Models are trained for 100 epochs, with learning rate $10^{-4}$ and batch size 4 using the MultiVae library \cite{senellart2025multivae}. 


\subsection{Data}

We use a dataset of 3D FDG PET scans from ADNI \cite{mueller2005ways} that are co-registered, averaged, and uniformized to the same resolution. We pre-process the images using the \texttt{pet-linear} pipeline from Clinica \cite{routier2021clinica}, which performs affine registration to the MNI space, cropping and intensity normalization using the average PET uptake in the cerebellum and pons regions. We select stable healthy patients that are constantly labeled as cognitively normal (CN) in ADNI for a 3-year interval. We use 545 images from 302 healthy subjects for training, and include two testing sets with 50 images from 50 subjects each, from CN (test-CN) and Alzheimer's disease (AD) subjects (test-AD). We split our training data in 5 folds to perform cross validation. Each fold contains roughly 450 images for training and 66 images for validation. Splits are performed at the subject-level and stratified by sex and age.

\subsection{Evaluation framework}
\label{sec:eval}

\textbf{Healthy-to-healthy reconstruction.} First we evaluate the ability of our model to reconstruct healthy images using classical metrics such as the mean squared error (MSE) and the structural similarity (SSIM) \cite{zhouwang2004image}. 
In what remains, we denote by $\text{hhMSE}$ the MSE between a healthy input $\x$ and its reconstruction $\hx$, i.e. $\text{MSE}(\hx, \x)$. 

\textbf{Simulation framework}
We also evaluate our model's ability to reconstruct pseudo-healthy images from abnormal samples.
Since ground truth pairs of abnormal and healthy images are not available, we use a simulation framework to simulate realistic hypometabolism characteristic of AD \cite{hassanaly2024evaluation}. We denote by $\xp = \x + \varepsilon$ the simulated abnormal version for a healthy image $\x$. We use two test datasets (test-AD-30 and test-AD-50), where we simulate AD with increasing severity.

\textbf{Abnormal-to-healthy reconstruction} We denote by $\text{ahMSE} = \text{MSE}(\hxp, \x)$, the $\text{MSE}$ between the pseudo-healthy reconstruction of $\xp$ and its ground truth healthy version $\x$. 

\textbf{Voxel-level anomaly detection metrics}
We design the anomaly maps as the residual $\mathbf{r} = \x - \hx$, normalized using a healthy validation set $\tilde{\mathbf{r}} = \frac{\mathbf{r} - \boldsymbol{\mu}_{HC}}{\boldsymbol{\sigma}_{HC}}$, where $\boldsymbol{\mu}_\text{HC}$ and $\boldsymbol{\sigma}_\text{HC}$ are the mean and standard deviation of $\mathbf{r}$ across all images in a healthy validation set (HC). 
To evaluate the unthresholded anomaly maps, we use their absolute value $|\tilde{\mathbf{r}}|$ to compute the voxel-wise average precision (AP) and \dice (best possible Dice index across thresholds) using the simulation mask as ground truth. 
Since these are sensitive to thresholding effects, we also define the $\text{rMSE} = \text{ahMSE} / \text{hhMSE}$ metric which measures reconstruction healthiness. We want this ratio to be close to 1, which corresponds to the case where the pseudo-healthy reconstruction from an abnormal image ($\text{ahMSE}$) is as good as the reconstruction from the healthy image ($\text{hhMSE}$).

\textbf{Sample-level evaluation on real data}
To evaluate our methods on real data, since ground truth segmentation masks are unavailable, we extract anomaly scores, defined as the mean of voxels of $|\tilde{\mathbf{r}}|$. 
We evaluate the unthresholded sample-level scores using sample-level AUC. 



\section{Trade-off between reconstruction quality and anomaly detection} \label{sec:trade-off}
\begin{figure}[htbp]
    \centering
    \includegraphics[width=\linewidth]{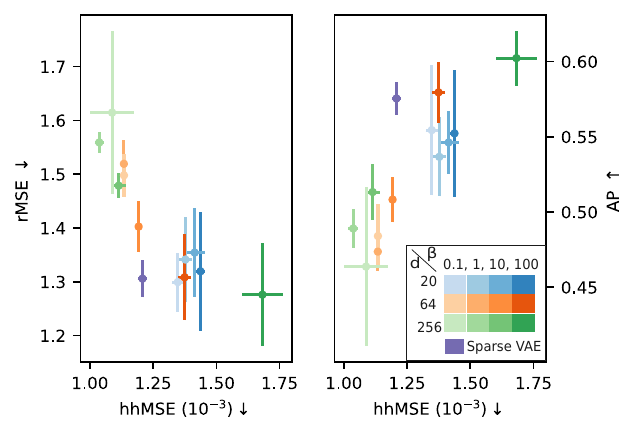}
    \caption{Trade-off between reconstruction quality (hhMSE) and anomaly detection performance (rMSE and AP) for $\beta$-VAEs with varying $\beta$ and latent dimension $d$. We report mean and standard deviation on five seeds for each model.}
    \label{fig:tradeoff}
\end{figure}

We explore the impact of constraining the latent space of $\beta$-VAEs through decreasing the latent dimension $d$ or increasing the $\beta$ weight. We train 12 models with $\beta \in [0.1,1.0,10,100]$ and $d \in [20,64,256]$. 
In \Cref{fig:tradeoff}, we report the anomaly detection performance (measured through
the MSE ratio
(rMSE = ahMSE / hhMSE)
and average precision) against the healthy reconstruction quality (hhMSE) for all models. 
We observe a reconstruction-detection trade-off among $\beta$-VAEs models. The less constrained the model (for instance low $\beta=0.1$ and high $d=256$), the better the reconstruction quality but the lower the detection performance. Overall, this analysis implies that the healthy reconstruction error (hhMSE) should not be the sole criterion for model selection in UAD. 
For the remaining of our experiments, we define as baseline the $\beta$-VAE with $d=64$ and $\beta=10$, which is a good compromise given the trade-off.

\section{Stability with respect to seeds}

For this baseline model ($d=64$, $\beta=10$), we study the variability of its performance across 30 random seeds and attempt to explain it by studying the distance between latent distributions of normal and abnormal images.
In particular, we model the latent codes' distributions for test-AD-30 and healthy test samples (test-CN) with multivariate Gaussian distributions and compare them with the 2-Wasserstein distance. To account for normal variability, we divide by the distance between the healthy training and validation distributions. 

\begin{figure}[htbp]
    \centering
    \includegraphics[width=0.65\linewidth]{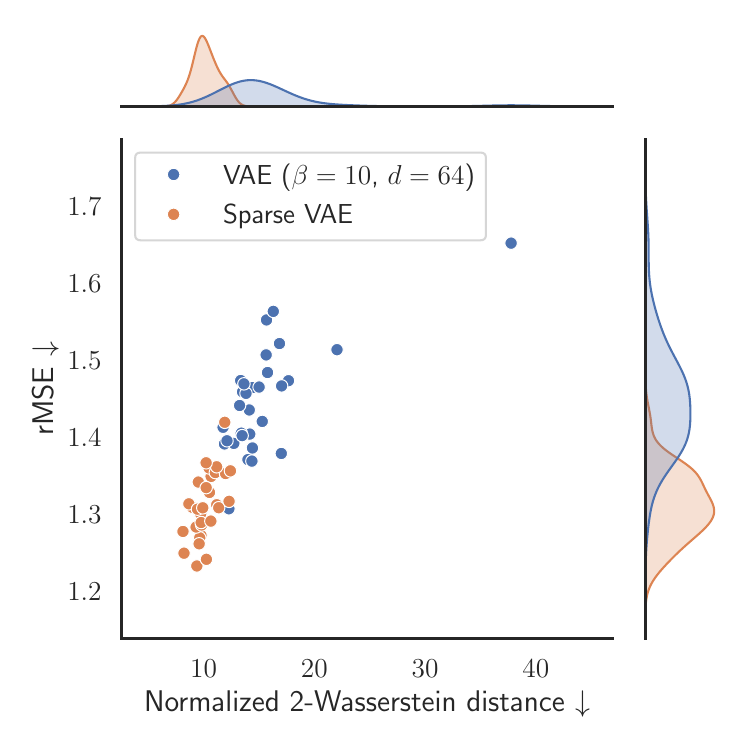}
    \caption{Scatterplot of the reconstruction healthiness (rMSE) on test-AD-30 against the normalized 2-Wasserstein distance. 
    Each point represents one of 30 seeds.}
    \label{fig:normmse_wasserstein}
\end{figure}

In \Cref{fig:normmse_wasserstein}, we plot the healthiness ratio (rMSE) against the normalized 2-Wasserstein distance between normal and abnormal samples for all 30 seeds.
By focusing on the rMSE distribution on the right, we can see that the model's performance is highly unstable when varying the seed.
From the scatterplot in Figure \ref{fig:normmse_wasserstein}, we observe that the distance between healthy and abnormal latent distributions is at least 10 times the distance between two healthy latent distributions. We also note that the model's performance (rMSE) correlates with the latent distance. This reveals that: (1) there is an important offset between the encoder outputs for normal and abnormal samples, and (2) this offset in the latent space is not compensated by the decoder since it directly impacts the reconstruction quality. 
Similar to what was observed by \cite{zhou2022rethinking}
at the sample level, best performing models are models that least separate normal and abnormal samples in the latent space. 
Overall, this analysis shows a detrimental lack of smoothness from the encoder and decoder functions. Let us consider a healthy $\x$ and abnormal $\xp = \x + \varepsilon$. If the encoder is not smooth enough, then $\mu(\xp)$ will be very different from $\mu(\x)$. If the decoder is also not smooth enough to compensate for this difference, then $\mathcal{D}(\mu(\xp))$ will again be very different from the optimal pseudo-healthy reconstruction $\mathcal{D}(\mu(\x))$.

\section{Building more robust VAE}
\label{sec:constrained_vae}

We explore two methods for improving the smoothness of the model and mitigate the trade-off observed in \Cref{sec:trade-off}.

\begin{table*}[htbp]
\centering
\scriptsize
    \begin{tabular}{l|ll|llll|llll}
        & \multicolumn{2}{c|}{test-CN} & \multicolumn{4}{c|}{test-AD-30} & \multicolumn{4}{c}{test-AD-50} \\
        & hhMSE \tiny{$10^{-3}$} $\downarrow$ & SSIM $\uparrow$ & ahMSE \tiny{$10^{-3}$} $\downarrow$ & rMSE $\downarrow$ & AP $\uparrow$ & $\lceil \text{Dice} \rceil$ $\uparrow$ & ahMSE \tiny{$10^{-3}$} $\downarrow$ & rMSE $\downarrow$ & AP $\uparrow$ & $\lceil \text{Dice} \rceil$ $\uparrow$ \\
        \midrule
        $\beta=1$ & $\mathbf{1.14 \pm 0.01}$ & $\mathbf{94.29 \pm 0.06}$ & $1.66 \pm 0.09$ & $1.48 \pm 0.07$ & $48 \pm 3$ & $49 \pm 2$ & $2.00 \pm 0.15$ & $1.78 \pm 0.13$ & $79 \pm 2$ & $72 \pm 2$ \\
        $\beta=10$ & $1.18 \pm 0.02$ & $94.20 \pm 0.05$ & $1.67 \pm 0.08$ & $1.42 \pm 0.09$ & $51 \pm 5$ & $51 \pm 3$ & $1.97 \pm 0.19$ & $1.68 \pm 0.17$ & $81 \pm 3$ & $73 \pm 3$ \\
        $\beta=30$ &$1.24 \pm 0.03$ & $94.11 \pm 0.12$ & $1.70 \pm 0.12$ & $1.38 \pm 0.07$ & $53 \pm 2$ & $52 \pm 2$ & $2.04 \pm 0.31$ & $1.67 \pm 0.22$ & $80 \pm 4$ & $73 \pm 4$ \\
        $\beta=50$ & $1.26 \pm 0.03$ & $94.07 \pm 0.07$ & $1.72 \pm 0.13$ & $1.37 \pm 0.07$ & $56 \pm 3$ & $54 \pm 2$ & $2.01 \pm 0.25$ & $1.60 \pm 0.16$ & $82 \pm 3$ & $74 \pm 2$ \\\midrule
        $\beta=30$ (cycle) & $1.20 \pm 0.02$ & $94.15 \pm 0.18$ & $1.61 \pm 0.07$ & $1.36 \pm 0.06$ & $56 \pm 3$ & $54 \pm 2$ & $1.89 \pm 0.16$ & $1.60 \pm 0.15$ & $83 \pm 2$ & $75 \pm 2$ \\
        $\beta=50$ (cycle) & $1.22 \pm 0.01$ & $94.22 \pm 0.06$ & $1.64 \pm 0.03$ & $1.35 \pm 0.03$ & $56 \pm 2$ & $54 \pm 2$ & $1.90 \pm 0.10$ & $1.57 \pm 0.09$ & $83 \pm 2$ & $75 \pm 2$ \\

        \midrule
        Sparse & $1.20 \pm 0.01$ & $94.21 \pm 0.02$ & $\mathbf{1.57 \pm 0.05}$ & $\mathbf{1.31 \pm 0.04}$ & $\mathbf{56 \pm 1}$ & $\mathbf{55 \pm 1}$ & $\mathbf{1.72 \pm 0.06}$ & $\mathbf{1.45 \pm 0.05}$ & $\mathbf{85 \pm 1}$ & $\mathbf{77 \pm 1}$ \\
        \bottomrule
    \end{tabular}
\caption{Reconstruction and detection results for healthy dataset (test-CN) and simulated AD datasets (test-AD-30 and test-AD-50). For each metric, we report the mean and standard deviation over 5 splits and seeds. Best metrics in bold. \emph{Erratum: the standard deviations in the ISBI 2026 proceedings were computed over the subjects and not splits.}}
\label{tab:results}
\end{table*}

\subsection{Cyclical beta-scheduling during training}

The simplest way to improve the smoothness of the decoder is to increase the $\beta$ weight on the KL regularization.
Indeed, for diagonal posteriors $q_{\phi}(z|x) = \mathcal{N}(\mu(x), \text{diag}(\sigma(x))$, the KL-divergence in \Cref{vae_loss} writes
\small
\begin{equation}
    \sum_{i=1}^d \frac{1}{2}\left({\sigma_i(x)}^2-\log (\sigma_i(x)^2) + \mu_i(x)^2\right)\,.
\end{equation}
\normalsize
For this regularization term to be small, we want $\forall i, \mu_i(x) \to 0$ and $\sigma_i(x)^2 \to 1$. Increasing $\beta$ encourages larger $\sigma_i(x)^2$, preventing posterior collapse (when $\sigma(x)^2 \to 0$). Larger $(\sigma_i(x)^2)_i$ forces the decoder function to be smoother, since during training, the decoder must reconstruct $x$ well from any $z\sim  \mathcal{N}(\mu(x), \text{diag}(\sigma(x))$. 

However, we have seen in \Cref{fig:tradeoff}, that increasing $\beta$ results in worse reconstruction quality. To counteract this effect, we vary the $\beta$ weight on the KL regularization, with a cyclical schedule during training as proposed in \cite{fu2019cyclical}. At the beginning of each cycle, $\beta=0$ in order to learn good reconstructions and it increases towards a maximum throughout the cycle to add structure to the latent space. We divide the training in 3 cycles and use large maximum values for $\beta \in \{30,50\}$ .

\subsection{Sparse VAE}

Another approach is the Sparse VAE model \cite{antelmi2019sparse} which implements gaussian dropout
for the latent codes during training.
In this model, the posterior distribution writes
\small
\begin{equation}
\label{eq:sparse-posterior}
    q_{\phi}(z|x) = \mathcal{N}\left(\mu(x), \text{diag}(\alpha\mu(x)^2)\right)\,,
\end{equation}
\normalsize
where $\alpha \in \mathbb{R}^d$ is a learned parameter that can be linked to a dropout ratio. The KL term rewrites such that it encourages $\alpha$ to be large. 
Unless a latent dimension $z_i$ is really useful for the reconstruction, $\alpha_i$ will become larger during training forcing the decoder to be less reliant on that dimension.
\Cref{eq:sparse-posterior} shows that the Sparse VAE also leads to larger encoder variances for large $|\mu(x)|$ creating a margin around the support of the healthy distribution on which the decoder can still reconstruct healthy images.
In our case, we empirically initialize $\forall i, \log\alpha_i= -3$, the highest value achieving a reconstruction error similar to our baseline $\beta$-VAE on healthy samples. 

\subsection{Results}

In \Cref{tab:results}, we report the results for all models on the simulated datasets. Each model is trained with $d=64$.
We observe that the $\beta$-scheduling improves the performance of $\beta$-VAE models: for $\beta=30$, the model with cyclical schedule reaches a lower hhMSE (1.20 vs 1.24) while reaching higher AP (56 vs 53) and \dice (54 vs 52).
Similarly, the Sparse VAE has a better abnormal-to-healthy MSE (1.57 vs 1.67) compared to the base model and also shows better anomaly detection results: the AP improves from 51 to 56 on test-AD-30. With more severe anomalies (test-AD-50) the Sparse VAE outperforms all other models on all metrics. Overall, this model shows an improvement over the trade-off of $\beta$-VAEs as displayed in
\Cref{fig:tradeoff}. 
In \Cref{fig:normmse_wasserstein}
, we further analyze this performance gain by showing that it comes with a smaller distance between normal and abnormal latent distributions. The marginal distributions in \Cref{fig:normmse_wasserstein} also show that this model is more stable across seeds. We also compute the standard deviation of the AP on 30 seeds, which is 2.0 for the Sparse VAE against 3.3 for the baseline model. 
Note that if we had selected a model based on the $\text{hhMSE}$ in \Cref{tab:results}, we would have chosen the $\beta=1$ model, which has the lowest detection performance, underlying the relevance of our analysis. 

\textbf{Results on real AD data}
We are interested in voxel-level anomaly detection but as a sanity check we
also compare the models on a test set of 50 CN and 50 real AD patients, by computing sample-level anomaly scores as described in \Cref{sec:eval}. All models achieve an AUC in the range of [80.6, 81] and none outperformed others in a significant way.

\section{Conclusion and future works}
\label{sec:conclusion}

We conduct an extensive empirical analysis of $\beta$-VAE models to highlight the trade-off between reconstruction quality and anomaly detection.
Second, we analyze the performance variability across different training seeds and find that it is linked with the distance between healthy and abnormal latent distributions. This confirms that there is a detrimental offset of abnormal latent codes compared to healthy ones that we aim to reduce. 
Based on these results, we explore two methods for overcoming these limitations: beta-scheduling and the Sparse VAE. While both methods 
allow to mitigate the trade-off by improving anomaly detection whilst maintaining good quality reconstructions, the Sparse VAE seems to reach slightly better performances and also proves more robust across different seeds.
To our knowledge, this is the first time that latent space adaptive regularization is explored to mitigate the reconstruction/detection trade-off at the voxel-level in medical imaging. 
Finally, the code used for obtaining the results presented is available on GitHub: \href{https://github.com/AgatheSenellart/tradeoff_vae_uad}{AgatheSenellart/tradeoff\_vae\_uad}. 

In future works, other methods could be explored such as the contractive autoencoder (CAE) \cite{rifai2011contractive} or sparse autoencoder (SAE) \cite{meng2017research} which impose different constraints such as L1 or L2 regularization on the weights or on the Jacobian of the encoder. These methods have yet to be tested for voxel-level anomaly detection. 
It would moreover be interesting to tackle other sources of variability such as the lack of purity of the training data. 
One could combine methods proposed here with the method proposed in \cite{akrami2022robust} modifying the loss to improve robustness to outliers in the training data.

\section{Acknowledgements}




This project was partly funded by the French government's Agence Nationale de la Recherche under the “France 2030” program (ANR-23-IACL-0008, PRAIRIE-PSAI), the ANO-NEURO project (ANR-23-CE45-0005-01), the ``Investissements d'avenir'' program (ANR-19-P3IA-0001); by the European Union’s Horizon Europe Framework Programme (grant 101136607, project CLARA) and ERC Synergy grant 101071601 (OCEAN). This work was performed using HPC resources from GENCI-IDRIS (grants AD011011648, AD011016015). Data collection and sharing for ADNI is funded by the National Institute on Aging (National Institutes of Health Grant U19AG024904).

\bibliographystyle{IEEEbib_initials}
\bibliography{refs}

\begin{thebibliography}{10}

\bibitem{chen2022unsupervised}
X.~Chen and E.~Konukoglu,
\newblock ``Unsupervised abnormality detection in medical images with deep
  generative methods,''
\newblock in {\em Biomedical {{Image Synthesis}} and {{Simulation}}}, pp.
  303--324. Elsevier, 2022.

\bibitem{baur2021autoencoders}
C.~Baur, S.~Denner, B.~Wiestler, N.~Navab, and S.~Albarqouni,
\newblock ``Autoencoders for unsupervised anomaly segmentation in brain {{MR}}
  images: {{A}} comparative study,''
\newblock {\em MedIA}, vol. 69, pp. 101952, 2021-04.

\bibitem{alaverdyan2020regularized}
Z.~Alaverdyan, J.~Jung, R.~Bouet, and C.~Lartizien,
\newblock ``Regularized siamese neural network for unsupervised outlier
  detection on brain multiparametric magnetic resonance imaging:
  {{Application}} to epilepsy lesion screening,''
\newblock {\em MedIA}, vol. 60, pp. 101618, 2020.

\bibitem{cai2024rethinking}
Y.~Cai, H.~Chen, and K.-T. Cheng,
\newblock ``Rethinking autoencoders for medical anomaly detection from a
  theoretical perspective,''
\newblock in {\em MICCAI}, 2024, vol. LNCS 15011, pp. 544--554.

\bibitem{zhou2022rethinking}
Y.~Zhou,
\newblock ``Rethinking {{Reconstruction Autoencoder-Based Out-of-Distribution
  Detection}},''
\newblock in {\em CVPR}. 2022, pp. 7369--7377, IEEE.

\bibitem{huijben2025enhancing}
E.~M.~C. Huijben, S.~Amirrajab, and J.~P.~W. Pluim,
\newblock ``Enhancing reconstruction-based out-of-distribution detection in
  brain {{MRI}} with model and metric ensembles,''
\newblock {\em Comp. Meth. and Prog. in Biomed.}, vol. 272, pp. 109045, 2025.

\bibitem{solal2026unsupervised}
M.~Solal, P.~André, and N.~Burgos,
\newblock ``Unsupervised anomaly detection in brain {{FDG PET}} with deep
  generative models: {{An}} experimental analysis of model variability and
  mitigation strategies,''
\newblock in {\em SPIE Medical Imaging}, 2026.

\bibitem{hassanaly2025benchmarking}
R.~Hassanaly, M.~Solal, O.~Colliot, N.~Burgos, and ADNI,
\newblock ``Benchmarking {{3D}} generative autoencoders for pseudo-healthy
  reconstruction of brain {{18F-FDG}} {{PET}},''
\newblock {\em JMI}, vol. 12, no. 05, 2025.

\bibitem{fu2019cyclical}
H.~Fu, C.~Li, X.~Liu, J.~Gao, A.~Celikyilmaz, and L.~Carin,
\newblock ``Cyclical annealing schedule: {{A}} simple approach to mitigating
  {{KL}} vanishing,''
\newblock in {\em NAACL}, 2019, pp. 240--250.

\bibitem{antelmi2019sparse}
L.~Antelmi, N.~Ayache, P.~Robert, and M.~Lorenzi,
\newblock ``Sparse multi-channel variational autoencoder for the joint analysis
  of heterogeneous data,''
\newblock in {\em ICML}. 2019, vol.~97, pp. 302--311, PMLR.

\bibitem{kingma2013autoencoding}
D.~P. Kingma and M.~Welling,
\newblock ``Auto-{{Encoding Variational Bayes}},'' 2013.

\bibitem{higgins2017betavae}
I.~Higgins, L.~Matthey, A.~Pal, C.~Burgess, X.~Glorot, M.~Botvinick,
  S.~Mohamed, and A.~Lerchner,
\newblock ``Beta-{{VAE}}: Learning basic visual concepts with a constrained
  variational framework,''
\newblock in {\em ICLR}, 2017.

\bibitem{senellart2025multivae}
A.~Senellart, C.~Chadebec, and S.~Allassonnière,
\newblock ``{{MultiVae}}: {{A Python}} package for multimodal variational
  autoencoders on partial datasets,''
\newblock {\em JOSS}, vol. 10, no. 110, pp. 7996, 2025.

\bibitem{mueller2005ways}
S.~G. Mueller, M.~W. Weiner, L.~J. Thal, R.~C. Petersen, C.~R. Jack, W.~Jagust,
  J.~Q. Trojanowski, A.~W. Toga, and L.~Beckett,
\newblock ``Ways toward an early diagnosis in {{Alzheimer}}'s disease: {{The
  Alzheimer}}'s {{Disease Neuroimaging Initiative}} ({{ADNI}}),''
\newblock {\em Alzheimer's \& Dementia}, vol. 1, no. 1, pp. 55--66, 2005.

\bibitem{routier2021clinica}
A.~Routier, N.~Burgos, (...), M.-O. Habert, S.~Durrleman, and O.~Colliot,
\newblock ``Clinica: An open-source software platform for reproducible clinical
  neuroscience studies,''
\newblock {\em Front. Neuroinform.}, vol. 15, pp. 689675, 2021.

\bibitem{zhouwang2004image}
{Zhou Wang}, A.~Bovik, H.~Sheikh, and E.~Simoncelli,
\newblock ``Image quality assessment: From error visibility to structural
  similarity,''
\newblock {\em IEEE Trans. on Image Process.}, vol. 13, no. 4, pp. 600--612,
  2004.

\bibitem{hassanaly2024evaluation}
R.~Hassanaly, C.~Brianceau, M.~Solal, O.~Colliot, and N.~Burgos,
\newblock ``Evaluation of pseudo-healthy image reconstruction for anomaly
  detection with deep generative models: {{Application}} to brain {{FDG
  PET}},''
\newblock {\em MELBA}, vol. 2, pp. 611--656, 2024.

\bibitem{rifai2011contractive}
S.~Rifai, P.~Vincent, X.~Muller, X.~Glorot, and Y.~Bengio,
\newblock ``Contractive auto-encoders: explicit invariance during feature
  extraction,''
\newblock in {\em ICML}. 2011, pp. 833--840, Omnipress.

\bibitem{meng2017research}
L.~Meng, S.~Ding, and Y.~Xue,
\newblock ``Research on denoising sparse autoencoder,''
\newblock {\em Int. J. Mach. Learn. \& Cyber.}, vol. 8, no. 5, pp. 1719--1729,
  2017.

\bibitem{akrami2022robust}
H.~Akrami, A.~A. Joshi, J.~Li, S.~Aydöre, and R.~M. Leahy,
\newblock ``A robust variational autoencoder using beta divergence,''
\newblock {\em Knowledge-Based Systems}, vol. 238, pp. 107886, 2022.

\end{thebibliography}

\end{document}